\documentclass[letterpaper, 10 pt, conference]{ieeeconf}  
\IEEEoverridecommandlockouts                              
\usepackage{xcolor}
\usepackage[pagebackref=false,breaklinks=true,letterpaper=true,colorlinks,bookmarks=false]{hyperref}
\usepackage{amsmath,graphicx}
\usepackage{algorithm}
\usepackage{algorithmic}
\usepackage{booktabs}
\usepackage{subcaption}
\usepackage{amssymb}
\usepackage[font=small]{caption}

\title{\LARGE \bf
\textbf{Sample-efficient Reinforcement Learning Representation Learning with \textbf{C}uriosity \textbf{C}ontrastive \textbf{F}orward \textbf{D}ynamics \textbf{M}odel}*
}

\author{Thanh Nguyen$^{\dagger}$, Tung M. Luu$^{\dagger}$, Thang Vu and Chang D. Yoo
\thanks{$^{\dagger}$ Equal contribution}%
\thanks{* This  work  was partly  supported  by  Institute  for  Information  \&communications  Technology  Planning  \&  Evaluation(IITP)grant funded by the Korea government(MSIT) (No. 2019-0-01396, Development of framework for analyzing, detecting,mitigating of bias in AI model and training data and No. 2021-0-01381, Development of Causal AI through Video Understanding and Reinforcement Learning)}%
\thanks{All authors are with Faculty of Electrical Engineering, Korea Advanced Institute of Science and Technology, Daejeon 34141, Republic of Korea {\tt\small Email: thanhnguyen@kaist.ac.kr}}%
}
\begin{document}

\maketitle
\thispagestyle{empty}
\pagestyle{empty}
\begin{abstract}
Developing an agent in   reinforcement learning (RL) that is capable of performing complex control tasks directly from high-dimensional observation such as raw pixels is a challenge as efforts still  need to be made towards improving sample efficiency and generalization of RL algorithm. This paper considers a learning framework for a Curiosity Contrastive Forward Dynamics Model (CCFDM) to achieve a more sample-efficient RL based directly on raw pixels. CCFDM incorporates a forward dynamics model (FDM) and performs contrastive learning to train its deep convolutional neural network-based image encoder (IE) to extract conducive spatial and temporal information to achieve  a more sample efficiency for RL. In addition, during training, CCFDM provides intrinsic rewards, produced based on FDM prediction error, and encourages the curiosity of the RL agent to improve exploration. The diverge  and less-repetitive observations provided by both our exploration strategy and data augmentation available in contrastive learning improve not only the sample efficiency but also the generalization . Performance of existing model-free RL methods such as Soft Actor-Critic built on top of CCFDM outperforms prior state-of-the-art pixel-based RL methods on the DeepMind Control Suite benchmark. 
\end{abstract}

\section{INTRODUCTION}
\begin{figure}[!t]
	\centering
	\vspace*{0.1cm}
	\includegraphics[width=\columnwidth]{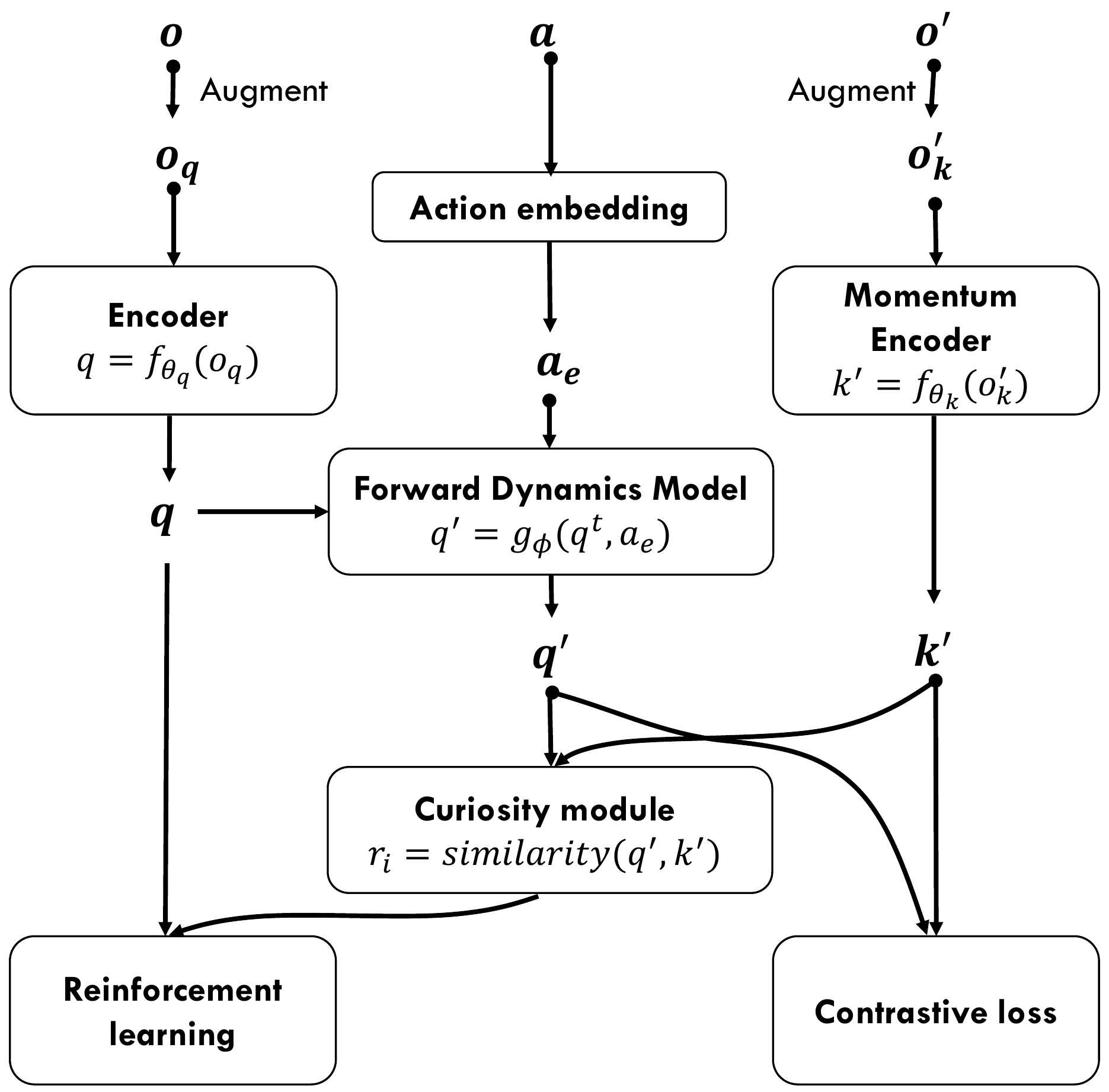}
	\caption{\textbf{C}uriosity \textbf{C}ontrastive \textbf{F}orward \textbf{D}ynamics \textbf{M}odel (CCFDM) uses data augmentation, contrastive learning and forward dynamics model (FDM) to enhance visual-based reinforcement learning (RL). CCFDM trains an image encoder (a.k.a query encoder-QE) by ensuring the extracted features of data-augmented observation of the current and next observation in the same transition are matched and far apart from other data-augmented observation features in other transitions using a contrastive loss. Here, $q$ is the query observation feature. Query $q'$ is generated by FDM from the query observation feature and action feature. Key $k'$ is generated by the momentum encoder (a moving average version of QE). The $k'$ consists of the positive key and negative keys. The query is forced to match with the positive key and far apart from the negative keys using contrastive learning objective. The RL algorithm is built on top of the query observation features and uses additional intrinsic reward signals from the curiosity module for better exploration. The whole framework is trained in an end-to-end manner.}
	\label{fig:conceptual}
\end{figure}

In recent years, Reinforcement Learning (RL) has received considerable attention for its achievements in games, robotics, and autonomous driving. In particular, the Deep Q-Network\cite{mnih2015human} outperforms human on Atari games \cite{bellemare13arcade}. The AlphaGo\cite{silver2016mastering} and AlphaZero \cite{silver2017mastering} defeat professional Go players. Recent advances in deep RL has even enabled agents to perform complex control tasks directly from visual observation of the environment: solving complex task from first-person view observations \cite{jaderberg2016reinforcement,espeholt2018impala}, autonomously performing robotic tasks   \cite{lillicrap2015continuous,levine2016end,lee2019stochastic,kalashnikov2018qt}.
	
Despite the aforementioned successes, sample efficiency and generalization are two main challenges in performing robotic tasks directly from visual observation. There has been remarkable progress in improving sample efficiency and generalization such as CURL \cite{srinivas2020curl}, RAD \cite{laskin2020reinforcement}, DrQ \cite{kostrikov2020image}, Planet \cite{hafner2019learning}, SAC-AE \cite{yarats2019improving}, SLAC \cite{lee2019stochastic}. Therein, combination of contrastive learning and data augmentation techniques from computer vision with model-free RL show certain improvements in sample efficiency on common RL benchmarks such as Atari \cite{bellemare13arcade}, DeepMind control \cite{tassa2020dmcontrol}, ProcGen \cite{cobbe2019procgen}, and OpenAI gym \cite{brockman2016openai}. However, these methods do not utilize temporal information of consecutive observations and strong exploration strategies, leading to limited performance.

This paper proposes \textbf{C}uriosity \textbf{C}ontrastive \textbf{F}orward \textbf{D}ynamics \textbf{M}odel (CCFDM), a framework that provides efficient image encoder learning for most of RL algorithms. CCFDM consists of an image encoder, a momentum encoder, a forward dynamics model (FDM), and a curiosity module. The connections of all modules are shown in Fig \ref{fig:conceptual}.
	
CCFDM wisely incorporates data augmentation and the FDM (a model that predicts next observation given current observation and action) to force the image encoder to capture both spatial and temporal information of visual observations by unsupervised contrastive learning. Consecutive observations potentially complement information each other or provide physics features. Image encoder has to utilize all information from provided transitions to extract meaningful features in order to satisfy both the FDM (temporal prediction) and the data augmentation (spatial disturbances) under CCFDM. In addition, contrastive learning helps extracted features more discriminative which is easier for training RL. 

 During training, the curiosity module provides intrinsic rewards in addition to extrinsic rewards from the environment. The intrinsic rewards encourage the curiosity of the RL agent to explore novel observations. The prediction error of the FDM is used as the intrinsic reward. Generating intrinsic rewards in this manner is efficient since it utilizes existing FDM and does not introduce extra models. Intuitively, the FDM prediction error (intrinsic reward) is high for novel observations, so that the agent is encouraged to explore these observations. Providing a more diverse and less repetitive set of observation not only makes the image encoder learning efficient but also improves generalization. 
	
The whole framework CCFDM is trained in an end-to-end manner with the  RL  algorithm is built on top of the query encoder and uses additional intrinsic reward signals from the curiosity module. CCFDM is evaluated on a diverse set of image-based continuous control tasks from DeepMind Control Suite \cite{tassa2020dmcontrol}. Empirical results showed that CCFDM improves feature representation learning in terms of data-efficiency and generalization, indicated by RL performance, compared to recent state-of-the-art methods.
	
The contributions of this paper are threefold: (i) incorporate FDM, augmentation, and contrastive learning to force the image encoder to capture meaningful features of visual observation, (ii) introduce simple but effective curiosity module which can utilize existing FDM to achieve better exploration, and (iii) provide an end-to-end framework which is compatible with most of RL algorithm and easy to implement. We thoroughly analyze the results and show that CCFDM improves feature representation learning in terms of data-efficiency and generalization compared to recent state-of-the-art methods.

\begin{figure*}
	\centering
	\includegraphics[width=\textwidth]{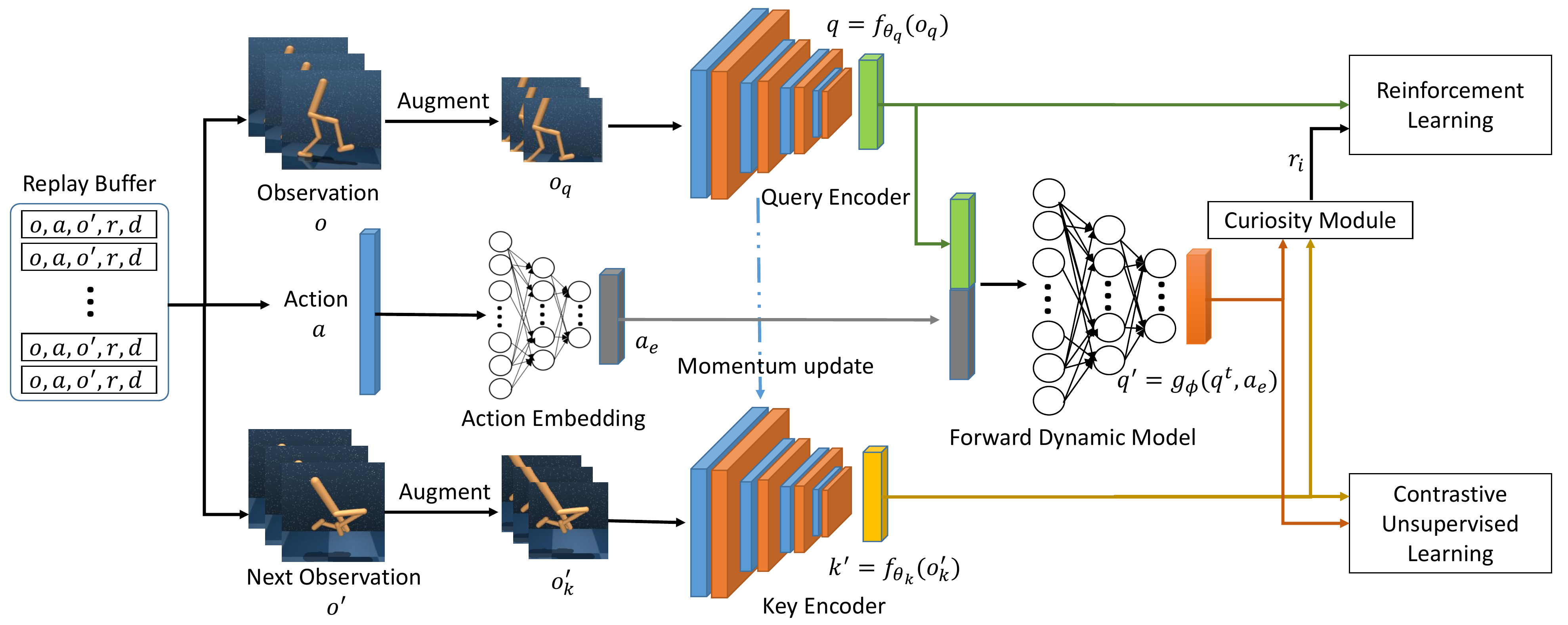}
	\small
	\caption{CCFDM Framework: A batch of transitions is sampled from the replay buffer. Observations are augmented to form query observations and next observations are augmented to form key observations. The query observations, the key observations,  and the actions are then encoded with the query encoder (image encoder), the key encoder, and the action embedding to form feature representations, respectively. Action feature ($a_e$) and query observation feature ($q$) are input to the forward dynamics model (FDM) to predict query ($q'$). The query-key pairs (the query, the positive key, and negative keys) are passed to the contrastive learning objective. The query-positive key pairs are passed to the curiosity module to calculate intrinsic rewards. The query observation features and the intrinsic rewards are passed to RL algorithm. During the gradient update, CCFDM updates the query encoder, the action embedding, the FDM simultaneously. The key encoder is the moving average version of the query encoder similar to MoCo \cite{he2020momentum}}.
	\label{fig:architecture}
\end{figure*}

\section{Related Work}

\begin{algorithm}[tb]
	\caption{Curiosity Contrastive Forward Dynamic Model}
	\small
	\label{alg:ccfdm_algorithm}
	\textbf{Input}:  Batch size $K$, training steps $ M $, EMA factor $ \tau$ , momentum update frequency $ \eta $ , intrinsic weight C, intrinsic decay weight $\gamma$
	
	\begin{algorithmic}[1] 
		\STATE \textbf{Initialize:} Replay Buffer $ \mathcal{D}$, QE $f_{\theta_q}$, KE $f_{\theta_k}$, FDM $ g_{\phi}$, AE $h_{\omega}  $
		\FOR {$ k = 1, M $}
		\STATE Sample transitions $ B = \{(o_j, a_j, o_j',r_j)\}_{j=1}^K $ from $ \mathcal{D}$
		\STATE Augment $B$ to get $ \hat{B} = \{(\hat{o}_j, a_j, \hat{o}_j',r_j)\}_{j=1}^K $
		\STATE Compute query observation feature, query, key: \\$ q = f_{\theta_q}(\hat{o}_j), q' = g_{\phi}(q,h_\omega(a)), k' = f_{\theta_k}(\hat{o}'_j) $
		\STATE Compute intrinsic reward $r_i$ by CM using Eq. \ref{eq:intrinsic_reward}
		\STATE Update $ f_{\theta_q} $, $ h_{\omega} $,  and $ g_{\phi} $ by contrastive loss in Eq. \ref{eq:infoNCE}
		\STATE Update $ f_{\theta_q} $, $ \mathcal{D}$ by RL algorithm with intrinsic rewards $r_i$ 
		\IF {$ k \mod \eta = 0$}
		\STATE Update momentum-based network $ f_{\theta_k} $:
		\STATE \hspace{1cm} $\theta_k \leftarrow \tau \theta_q + (1 - \tau)\theta_k$
		\ENDIF
		\ENDFOR
	\end{algorithmic}
\end{algorithm}

\textbf{Reinforcement Learning (RL)}. RL is a research field in machine learning that aims to form a software agent that can perform actions in an environment so as to maximize some notion of cumulative reward. RL algorithms can be classified into model-based RL (e.g: \cite{racaniere2017imagination, ha2018recurrent,feinberg2018model,nagabandi2018neural}) and model-free RL (e.g: \cite{sorokin2015deep, li2019robust}). Model-based RL has access to (or learns) a model of the environment and it provides better sample efficiency than model-free RL. In case model-based RL has to learn the model of the environment, it is extremely difficult to achieve performance better than that of model-free RL: the learned model is usually less accurate than that of model-free RL. By contrast, model-free RL foregoes the potential gains in sample efficiency for gaining convenience in design.  These algorithms outperform humans at board games \cite{silver2016mastering,silver2017mastering}, computer games \cite{schrittwieser2020mastering,vinyals2019grandmaster}, and complex robotic tasks \cite{openai2019solving,andrychowicz2017hindsight}. CCFDM can be used for most model-free reinforcement learning (RL).

\textbf{Intrinsic Reward exploration}. Intrinsic rewards as exploration bonuses are one of the well-known approaches to better exploration, especially for solving hard-exploration problems \cite{bellemare2016unifying}. Intrinsic reward exploration is also known as curiosity exploration and the methods are diverge ranging from count-based exploration (counting by density model \cite{bellemare2016unifying,ostrovski2017count,zhao2019curiosity}, counting after hashing \cite {tang2017exploration}) to prediction-based exploration (forward dynamics \cite{oudeyer2007intrinsic}, \cite{pathak2017curiosity,burda2018large} , random networks \cite{choshen2018dora,burda2018exploration}, physical properties \cite{denil2016learning}). Intrinsic reward exploration has improved sample efficiency in conventional RL. CCFDM utilizes prediction-based exploration using a forward dynamics model and incorporates it to the framework of contrastive learning for visual observation RL.

\textbf{Dynamics Models for Sample-efficient RL}. Modeling the dynamics of the environment has proved effective in improving sample efficiency. The dynamics model can be used for generating more data for Atari games \cite{kaiser2019model}, planing ahead  \cite{hafner2019learning,hafner2019dream}, shaping the representations using an auxiliary loss \cite{jaderberg2016reinforcement,oord2018representation,lee2019stochastic}. CCFDM estimates a forward dynamics model to force the image encoder to capture temporal information and learn meaningful representation which is sufficient for dynamics prediction. 

\textbf{Representation Learning for RL}. Representation learning plays an important role in achieving high performance in visual-based RL. Many methods have been proposed with various approaches.	\cite{jaderberg2016reinforcement} proposes representation auxiliary loss for improving RL performance in the DeepMind Control Suite \cite{beattie2016deepmind}. Using a reconstruction-based task is also an effective approach. Beta variation autoencoder ($ \beta\text{-VAE} $) \cite{kingma2013auto} or deterministic autoencoder AE is leveraged for encoding features. \cite{higgins2017darla,yarats2019improving} jointly learn VAE/AE objectives and RL objectives while \cite{finn2015learning,nair2018visual} propose to train the two objectives in an alternating fashion. Recently, contrastive learning in computer vision \cite{chen2020improved,he2020momentum,chen2020simple} are leveraged to improve representation learning for RL algorithms. \cite{oord2018representation} proposed a variant of the noise-contrastive estimation loss on future steps on top of the base model A2C \cite{mnih2016asynchronous}. \cite{anand2019unsupervised} introduces a new contrastive loss to improve sample efficiency in Atari benchmark. \cite{srinivas2020curl} proposed a general framework that is a combination of contrastive loss with image augmentation for learning the features. \cite{laskin2020reinforcement,kostrikov2020image} proved that data augmentation can significantly improve sample efficiency for learning directly from visual observation. This paper incorporates contrastive learning, data augmentation, and FDM in one framework for improving both sample efficiency and generalization in visual-based RL. 
	
	\begin{figure*}[t]
		\centering
		\begin{subfigure}{1.0\textwidth}
			\centering
			\begin{subfigure}{0.16\columnwidth}
				\centering
				\includegraphics[width=0.97\textwidth]{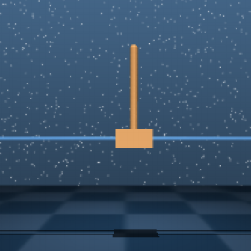}
				\caption{Cartpole}
				\label{fig:input}
			\end{subfigure}
			\begin{subfigure}{0.16\columnwidth}
				\centering
				\includegraphics[width=0.97\textwidth]{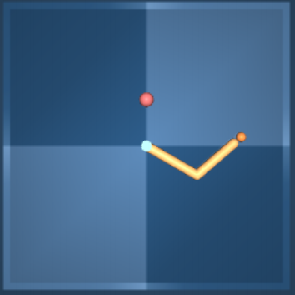}
				\caption{Reacher}
			\end{subfigure}
			\begin{subfigure}{0.16\columnwidth}
				\centering
				\includegraphics[width=0.97\textwidth]{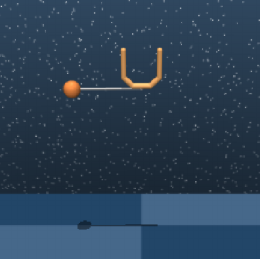}
				\caption{Ball in cup}
			\end{subfigure}
			\begin{subfigure}{0.16\columnwidth}
				\centering
				\includegraphics[width=0.97\textwidth]{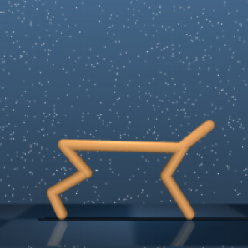}
				\caption{Cheetah}
			\end{subfigure}
			\begin{subfigure}{0.16\columnwidth}
				\centering
				\includegraphics[width=0.97\textwidth]{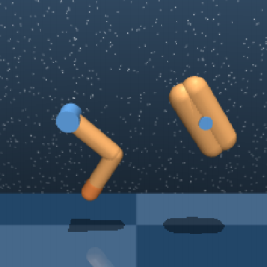}
				\caption{Finger}
			\end{subfigure}
			\begin{subfigure}{0.16\columnwidth}
				\centering
				\includegraphics[width=0.97\textwidth]{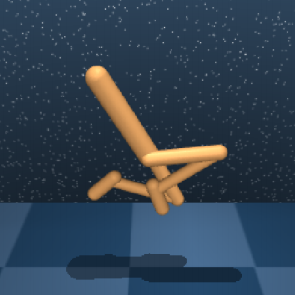}
				\caption{Walker}
				\label{fig:instance_pred}
			\end{subfigure}
		\end{subfigure}
        \caption{CCFMD is benchmarked on image-based continuous control tasks from the DeepMind Control Suite (DMC) \cite{tassa2018deepmind}. DMC offers six excellent RL domains (physical model environments) which introduce challenges of sparse reward (e.g: ball in cup, finger, reacher), dense reward (e.g: cart pole, cheetah, walker), complex dynamics (e.g: finger, cheetah), hard exploration (e.g: walker) and other traits. Each domain consist of smaller tasks (e.g: walker walk, walker stand, walker run)}
		\label{fig:DMCbench}
\end{figure*}	
\section{Background}

\subsection{Reinforcement Learning with Intrinsic Reward}
Reinforcement Learning (RL) considers solving a Markov Decision Process (MDP). MDP is characterized by a set of state $\mathcal{O}$, a set of action $\mathcal{A}$, a transition probability $P$ mapping from current observation $o$ and action $a$ to the future observation $o'$, and the immediate reward $R$, also known as extrinsic reward (denoted by $r_e$).  Given the policy  $\pi \in \Pi$: $O \rightarrow A$, the goal of RL is to learn an optimal policy that maximizes the expected cumulative reward:

\begin{align}
F(\pi)=\mathbb{E}_{a_t \sim \pi}\left[ \sum_{t=0}^{\infty} \gamma^{t} r_e (o_{t}, a_{t}) \right].
\end{align}

RL algorithms that explicitly estimate (or are given) transition probability are considered model-based RL, the others are considered model-free RL. 

To obtain a good policy, the RL algorithm has to balance between exploration (discover novel observations) and exploitation (follow the current best policy). An effective approach is to provide exploration bonuses as rewards (a.k.a. intrinsic reward $r_i$) to encourage policy reaching novel observation during training in order to provide a diverse set of observations. This prevents the policy from getting stuck in the local minimal. Thus, during training, the policy has to maximize the new expected cumulative reward:

\begin{align}
   F(\pi)=\mathbb{E}\left[ \sum_{t=0}^{\infty} \gamma^{t} (r_e\left(o_{t}, a_{t}\right) +r_i\left(o_{t}, a_{t}\right))\right]. 
\end{align}

\subsection{Soft Actor-Critic}
Soft Actor-Critic (SAC) \cite{haarnoja2018soft} is an effective RL algorithm for learning robotics tasks. SAC learns a policy $\pi_\theta$ (a.k.a actor) and critics $Q_{\phi_1}$ and $Q_{\phi_2}$. $\phi_i$ are learning by minimizing the Bellman error:
\begin{align}
    \mathbb{E}_{b \sim \mathcal{D}}\left[\left(Q_{\phi_{i}}(o, a)-(r_e+\gamma(1-d) \mathcal{T})\right)^{2}\right],
\end{align}
where $b =(o,a,o',r_e,d)$, $d$ is done signal, $\mathcal{D}$ is the replay buffer, and $\mathcal{T}$ is defined as:
\begin{align}
    \mathcal{T} = \min _{i=1,2} \left[ Q_{\phi_{i}}^{*}\left(o^{\prime}, a^{\prime}\right)-\alpha \log \pi_{\theta}\left(a^{\prime} \mid o^{\prime}\right)\right].
\end{align}
The $Q_{\phi_{i}}^{*}$ denotes the exponential moving average (EMA) of the parameters of $Q_{\phi_{i}}$ and $\alpha$ is a positive entropy coefficient. The actor is trained by maximizing the expected return of its actions as in: 
\begin{align}
    \mathcal{L}(\theta)=\mathbb{E}_{a \sim \pi}\left[Q^{\pi}(o, a)-\alpha \log \pi_{\theta}(a \mid o)\right].
\end{align}

\subsection{Contrastive Learning}

Contrastive learning is an approach to improve representation learning by a teaching model whereby pairs of data points are “similar” or “different”. To be specific, given a query q and keys $K = \{k0, k1, . . . \}$ where K includes the positive key $k+$ and the negative keys $K\backslash\{k+\}$. The goal of contrastive learning is to ensure that q matches with $k+$ and is far apart from $K\backslash \{k+\}$. One example of contrastive learning loss \cite{he2020momentum} is as below:

\begin{align}
\mathcal{L}_{q}=-\log \frac{\exp \left(q \cdot k^{+} / \tau\right)}{\sum_{i=0}^{K} \exp \left(q \cdot k_{i} / \tau\right)}
\end{align}

where $\tau$  is a temperature hyper-parameter.

\section{CCFDM Implementation}

The proposed framework extends the model-free RL algorithm by adding modules to aid image encoder parts. The whole architecture of the framework is shown in Fig \ref{fig:architecture}. There are five main modules: query encoder (QE), key encoder (KE), action embedding (AE), forward dynamics model (FDM), and curiosity module (CM). The QE is a deep convolutional neural network that maps visual observation to a feature vector. The KE is the moving average version (EMA) of the QE similar to MoCo \cite{he2020momentum} which helps stabilize the learning process of QE. The AE and the FDM is multi-layer perceptron networks. Therein, AE encodes action to form an appropriate action feature for FDM to predict the next observation feature (a.k.a query). The curiosity module is a non-learning module that provides intrinsic rewards.

Formally, we denote the QE is $f_{\theta_q}$ the KE is $f_{\theta_k}$, the AE is $h_\omega$, the FDM is $g_\phi$ and replay buffer is $\mathcal{D}$. A work flow of framework is as follow. A batch of transitions $\mathcal{B}$ are sampled from $\mathcal{D}$. Observations are augmented to form query observations $o_q$ and next observations are augmented to form key observations $o_k'$. The query observations, the key observations,  and the actions $a$ are then encoded with the QE, the KE, and the AE to form feature representations: 
\begin{align}
q=f_{\theta_q(o_q)} ;k'=f_{\theta_k(o'_k)}; a_e = h_\omega(a)
\end{align}
Action feature ($a_e$) and query observation feature ($q$) are input to FDM to predict query ($q'$): $q' =g_\phi(q,a_e)$. The query-key pairs (the query, the positive key, negative keys) are passed to the contrastive learning objective. Any contrastive learning objective can be used but we found that infoNCE \cite{wu2018unsupervised} showed the best performance. The formulation is below:

	\begin{equation}\label{eq:infoNCE}
	\resizebox{.90\linewidth}{!}{$
		\displaystyle
		\mathcal{L}(q',k') = - \log\frac{\exp{(\text{sim}(q', k'^{+})})}{\exp{(\text{sim}(q', k'^+)}) + \sum_{j=1}^{K-1}\exp{(\text{sim}(q', k'_j))}}
		$},
	\end{equation}%
Where j is the sample index in $\mathcal{B}$, $K$ is the batch size and $\text{sim}(\cdot)$ is the similarity measure function. It can be Bilinear products $\text{sim}(q',k') = (q'^TWk')$ or dot products $\text{sim}(q',k') = (q'^Tk')$.  

\begin{table*}[hbt!]
        \footnotesize
		\vskip 0.1in
		\centering
		\begin{center}
				\begin{sc}
				\begin{tabular}{lccccccc|c}
					\toprule
					500K STEP SCORES & CCFDM (Ours) &DrQ   & SAC-CURL 		       & PlaNet   & SAC-AE  & SLAC   & SAC-Pixel	& SAC State\\ 
					\midrule
                    Finger,spin       & 906$\pm$152         & \textbf{938$\pm$103} & 874$\pm$151 & 418$\pm$382 & 884$\pm$128 & 771$\pm$203 & 179$\pm$166 & 927$\pm$43 \\
                    Cartpole,swingup  & \textbf{875$\pm$38} & 868$\pm$10           & 861$\pm$30           & 464$\pm$50  & 735$\pm$63  & -       & 419$\pm$40  & 870$\pm$7  \\
                    Reacher,easy      & \textbf{973$\pm$36} & 942$\pm$71           & 904$\pm$94           & 351$\pm$483 & 627$\pm$58  & -       & 145$\pm$30  & 975$\pm$5  \\
                    Cheetah,run       & 552$\pm$130         & \textbf{660$\pm$96}  & 500$\pm$91  & 321$\pm$104 & 550$\pm$34  & 629$\pm$74  & 197$\pm$15  & 772$\pm$60 \\
                    Walker,walk       & \textbf{929$\pm$68} & 921$\pm$46           & 681$\pm$68           & 293$\pm$114 & 847$\pm$48  & 865$\pm$97  & 42$\pm$12   & 964$\pm$8  \\
                    Ball in cup,catch & \textbf{979$\pm$17} & 963$\pm$9            & 958$\pm$13           & 352$\pm$467 & 794$\pm$58  & 959$\pm$4   & 312$\pm$63  & 979$\pm$6 \\
					\midrule
					100K STEP SCORES  & & &  &  &  & &  &\\
					\midrule
                    Finger,spin       & 880$\pm$142          & \textbf{901$\pm$104} & 779$\pm$108 & 95$\pm$164  & 740$\pm$64 & 680$\pm$130 & 179$\pm$66 & 672$\pm$76  \\
                    Cartpole,swingup  & \textbf{785$\pm$87}  & 759$\pm$92           & 592$\pm$170          & 303$\pm$71  & 311$\pm$11 & -       & 419$\pm$40 & 812$\pm$45  \\
                    Reacher,easy      & \textbf{811$\pm$220} & 601$\pm$213          & 517$\pm$113          & 140$\pm$256 & 274$\pm$14 & -       & 145$\pm$30 & 919$\pm$123 \\
                    Cheetah,run       & 274$\pm$98           & 344$\pm$67  & 307$\pm$48  & 165$\pm$123 & 267$\pm$24 & \textbf{391$\pm$47}  & 197$\pm$15 & 228$\pm$95  \\
                    Walker,walk       & \textbf{634$\pm$132} & 612$\pm$164          & 323$\pm$43           & 125$\pm$57  & 394$\pm$22 & 428$\pm$74  & 42$\pm$12  & 604$\pm$317 \\
                    Ball in cup,catch & \textbf{962$\pm$28}  & 913$\pm$53           & 772$\pm$241          & 198$\pm$442 & 391$\pm$82 & 607$\pm$173 & 312$\pm$63 & 957$\pm$26  \\
					\bottomrule
				\end{tabular}
				\end{sc}
		\end{center}
		\caption{Scores achieved by CCFDM (mean \& standard deviation) and baselines on DMC evaluated at 500k environment step and 100k environment step. CCFDM achieves state-of-the-art performance on the 4 out of 6 environment and just below  DRQ on Finger-Spin and Cheetah-Run. The baseline are DrQ \cite{kostrikov2020image}, SAC-CURL \cite{srinivas2020curl},PlaNet \cite{hafner2019learning}, SAC-AE \cite{yarats2019improving}, SLAC \cite{lee2019stochastic}, SAC-Pixel	and  SAC State \cite{haarnoja2018soft}.}
		\vskip -0.1in
		\label{table:results_dmcontrol}
	\end{table*}

The query-positive key pairs are passed to the curiosity module to calculate intrinsic rewards $r_i$. The curiosity module measures the similarity between the query and positive key then gives intrinsic reward proportional to the dis-similarity. The similarity is not bounded and differs from task to task. We normalize the similarity to make it task-agnostic. Furthermore, the intrinsic reward is decayed during training to make agents converge to the optimal solutions. The formulation for CM is below:

\begin{equation}
	r_i(q,k) = Ce^{(-\gamma t)}\text{sim}(q,k)\frac{r_{e}^{max}}{r_i^{max}}
	\label{eq:intrinsic_reward}
\end{equation}%
	
Therein, $t$ is environment step, $C$ is temperature weight, $\gamma$ is decay weight, $r_{e}^{max}$ and $r_i^{max}$ are the maximum extrinsic reward value and intrinsic reward value over t, respectively.  

Finally, the $q$ and $r_i$ , $\mathcal{B}$ are passed to the RL algorithm. During the gradient update, the CCFDM updates the query encoder, action embedding, and FDM simultaneously. Note that $\mathcal{D}$ is updated during RL training. The whole algorithm is described the Alg. \ref{alg:ccfdm_algorithm} 
\section{Experiment and result}
\label{sec:typestyle}

\subsection{Experiment setup}

\begin{figure}[htp]
	\centering
	\vspace*{0.1cm}
	\includegraphics[width=\columnwidth]{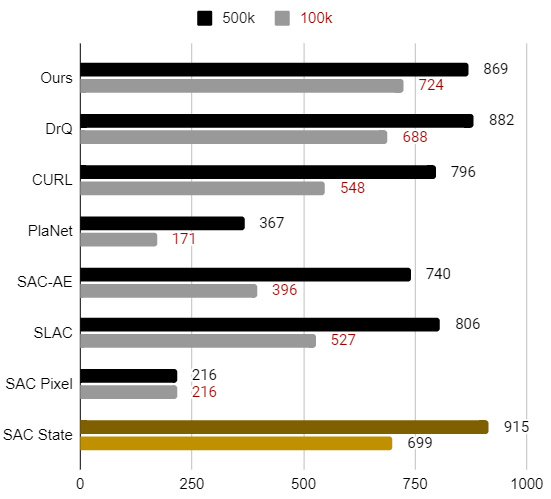}
	\caption{Evaluation Score Performance of CCFDM averaged over six tasks relative to DrQ, CURL, PlaNet, SAC-AE, SLAC, SAC Pixel, and SAC state. CCFDM out-performs all the baselines and nearly reaches the SAC state which is considered the upper bound performance.}
\label{fig:averagereturn}
\end{figure}

The proposed framework is benchmarked on six continuous control tasks from DeepMind Control suite (DMC) \cite{tassa2020dmcontrol}. DMC is considered as a standard benchmark for evaluating visual observation RL algorithms \cite{srinivas2020curl,laskin2020reinforcement,kostrikov2020image} in term of sample efficiency and generalization. DMC provides excellent physical model environments referred to as domains. Each domain has different tasks associated with a particular MDP structure. Our experiments are conducted on six well-known domains as shown in Figure \ref{fig:DMCbench}. The following tasks were chosen: Cartpole-Swingup, Ball in cup-Catch, Reacher-Easy, Finger-Spin, Cheetah-Run, and Walker-Walk since they provide a diverse set of challenges including sparse reward, dense reward, complex dynamics, hard exploration, and other traits \cite{yarats2019improving}. The specific task setup is as follows \cite{hafner2019learning,laskin2020reinforcement,srinivas2020curl} for fair comparison.

\begin{figure*}[htp]
	\centering
	\begin{subfigure}{0.32\linewidth}
	\centering
		\includegraphics[width=\textwidth]{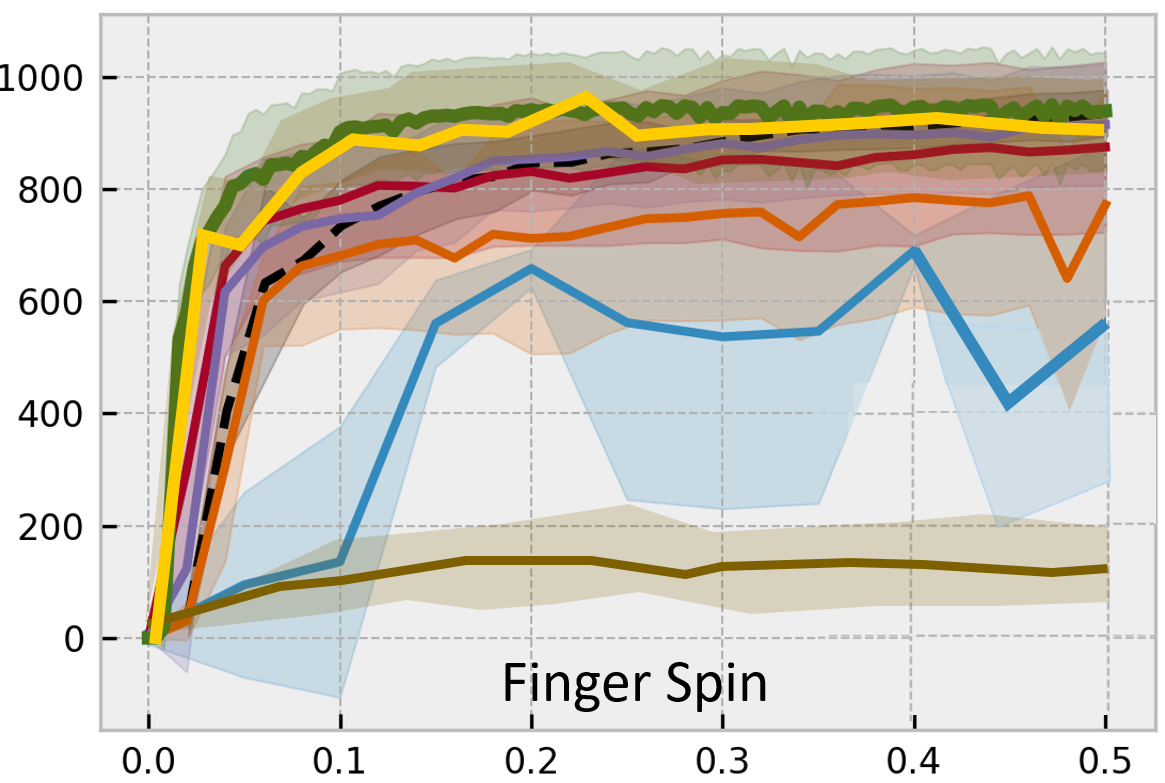}
	\end{subfigure}
	\begin{subfigure}{0.32\linewidth}
	\centering
		\includegraphics[width=\textwidth]{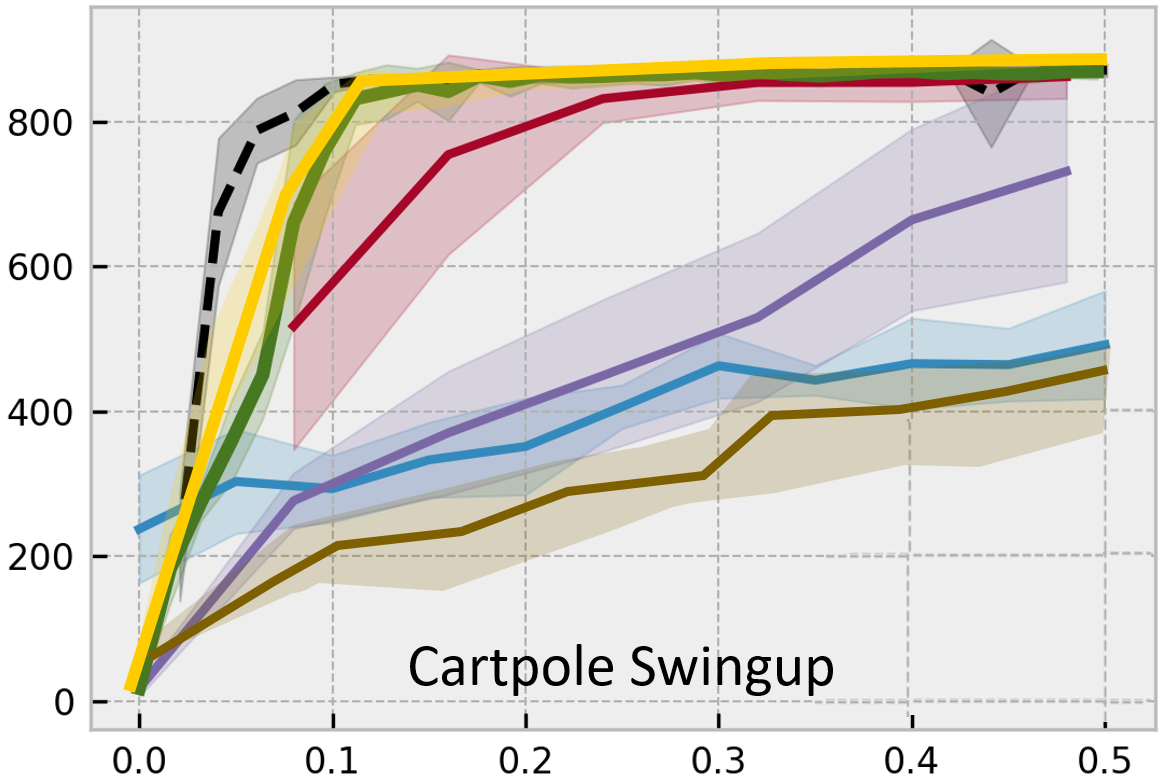}
	\end{subfigure}
	\begin{subfigure}{0.32\linewidth}
	\centering
		\includegraphics[width=\textwidth]{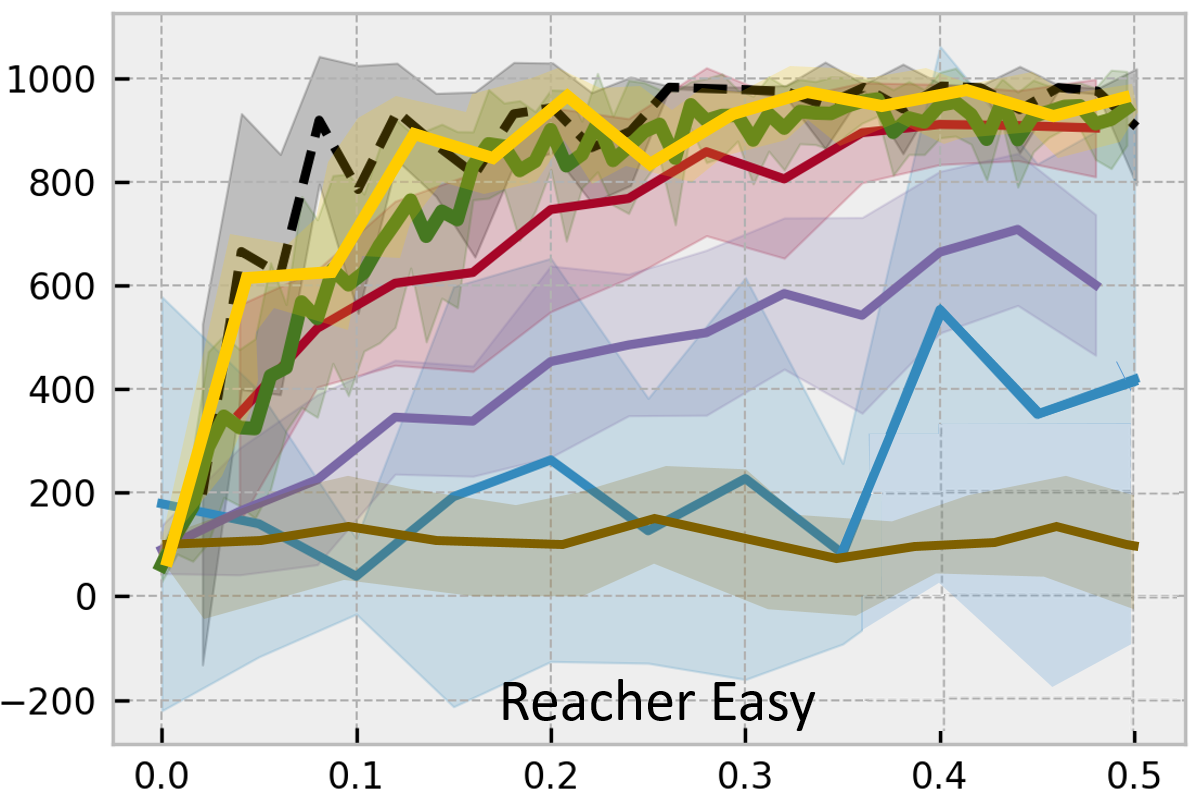}
	\end{subfigure}
	
	\begin{subfigure}{0.32\linewidth}
	\centering
		\includegraphics[width=\textwidth]{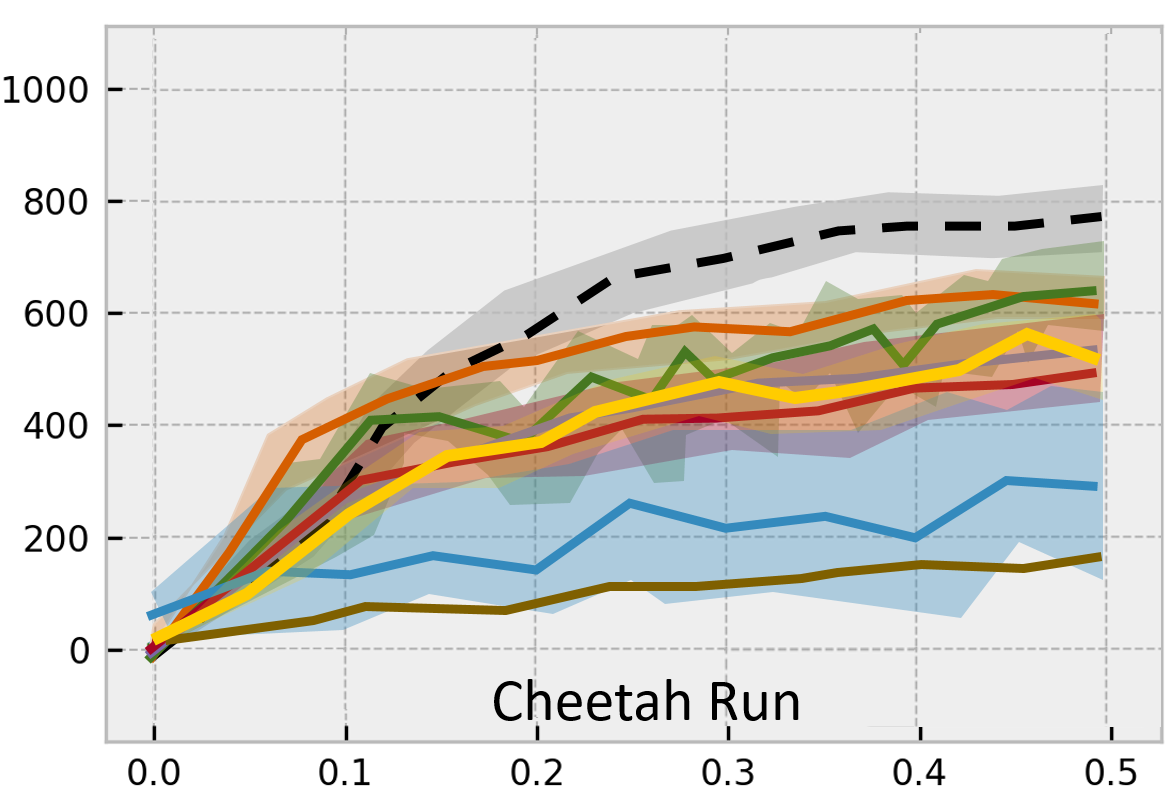}
	\end{subfigure}
	\begin{subfigure}{0.32\linewidth}
	\centering
		\includegraphics[width=\textwidth]{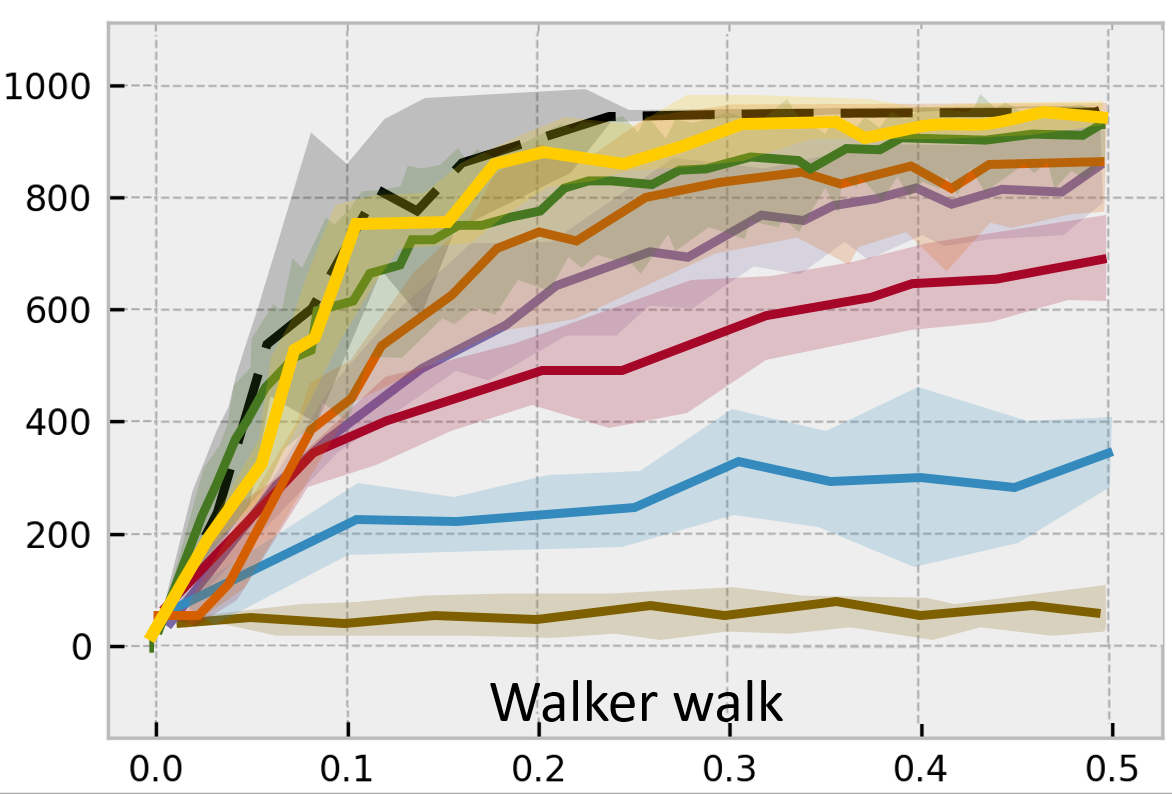}
	\end{subfigure}
	\begin{subfigure}{0.32\linewidth}
	\centering
		\includegraphics[width=\textwidth]{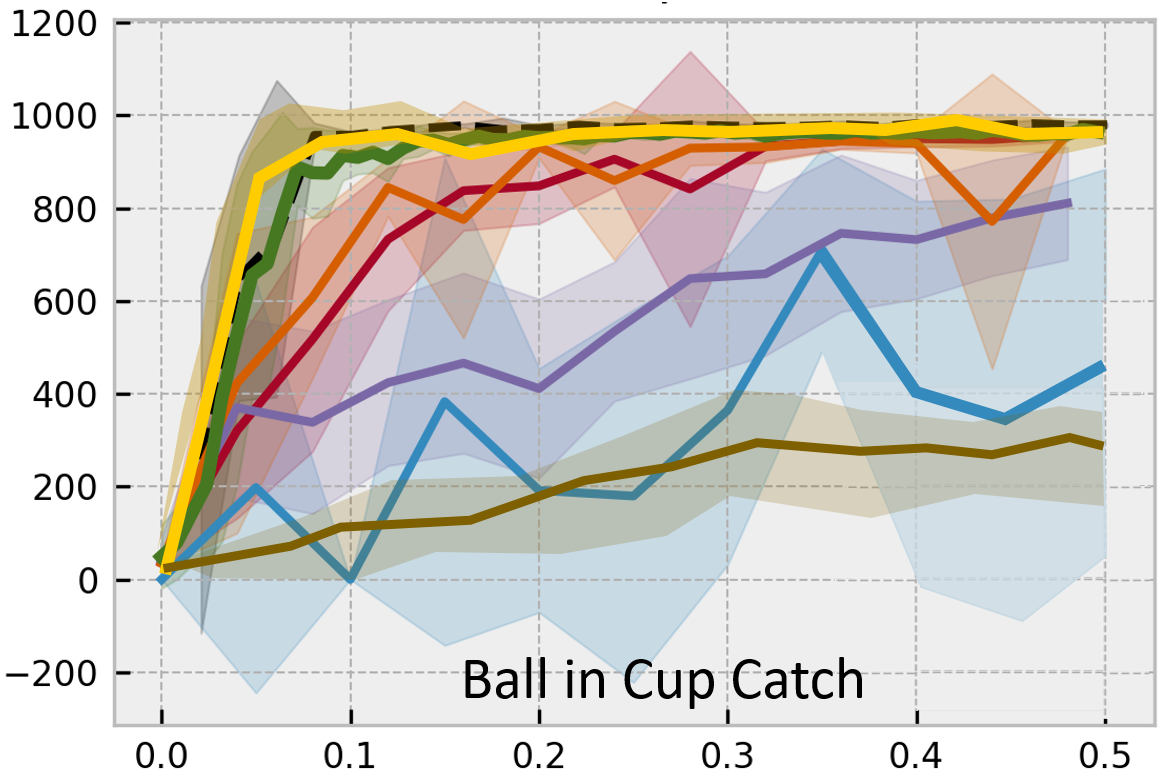}
	\end{subfigure}
	
	\begin{subfigure}{0.5\linewidth}
	\centering
		\includegraphics[width=\textwidth]{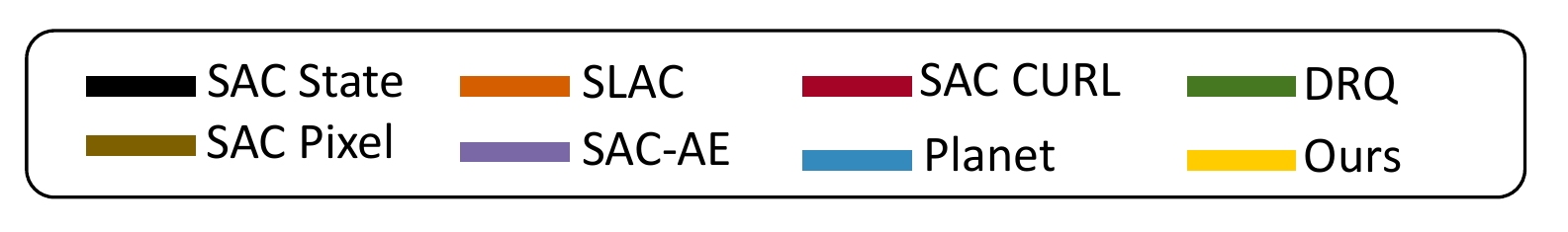}
	\end{subfigure}
	\caption{Comparison of the data-efficiency among recent state-of-the-art methods indicated by average evaluation score during the whole training process \protect\footnotemark. CCFDM matchs state-of-the-art performance of DrQ and significantly out-performs other baselines in five over six environment.}
	\label{fig:sampleEfficiency}
\end{figure*}

CCFDM uses the same network model architectures for all tasks to evaluate the robustness of the framework even though carefully choosing network model architectures potentially generates better results. In particular, the query encoder (QE) architecture consists of four convolutional layers with \texttt{ReLU} activation followed by a fully connected projection layer which is similar to SAC-AE \cite{yarats2019improving}. The forward dynamics model (FDM) and action embedding (AE) are modeled as Multi-Layer Perceptron with two hidden layers of 50 \texttt{ReLU} units. The key encoder (KE) architecture is identical to QE. The KE weights are the moving average of the KE weights which are similar to MoCo \cite{he2020momentum}. The EMA coefficient $ \tau = 0.01 $. Soft Actor Critic (SAC) \cite{haarnoja2018soft} is used as a base RL algorithm similar to \cite{laskin2020reinforcement,yarats2019improving}. The actor and critic use the QE for feature extraction. As with previous algorithms \cite{yarats2019improving,laskin2020reinforcement}, the batch size is set to 512, the target critic and the target actor are updated every two updates of the main critic. We use \textit{random cropping} for data augmentation throughout the experiments. The FDM is optimized using Adam optimizer \cite{kingma2013auto} with default parameters and initial learning rate $ 1e-3 $. Intrinsic reward decay weight is set to 2e-5 and intrinsic weight is set to 0.2. All other settings are the same as mentioned in Curl \cite{srinivas2020curl}.

\subsection{Experiment result}

During training, we simultaneously evaluated the RL agent every 10K environment steps with 10 episodes and logged the average test returns. For each task, we trained our algorithm five times with different seeds and report the result as Table \ref{table:results_dmcontrol}. The result demonstrates CCFMD significantly improved performance over the baselines: SAC-CURL \cite{srinivas2020curl}, PlaNet \cite{hafner2019learning}, SAC-AE \cite{yarats2019improving}, SAC-Pixel in all tasks.  CCFDM achieves state-of-the-art performance on four out of six tasks and just below DrQ  on  the Finger-Spin  and the Cheetah-Run. Moreover, CCFDM nearly reaches the performance of learning from the state indicated by performance of the SAC State \cite{haarnoja2018soft}. 

The sample-efficiency is evaluated by the performance at 100k environment step and 500k environment step as proposed in CURL \cite{srinivas2020curl}. The average result over six tasks is shown in Fig \ref{fig:averagereturn}. According to the result, at the $ 100K $ environment step, CCFDM gains 3.3$\times $ higher median performance than learning from pixel only (SAC Pixel), around 1.3$\times $ higher than SLAC and CURL. CCFDM converges close to an optimal score of 1000 on all six tasks within 500k steps. It also nearly matches the SAC State together with the DrQ. The Fig  \ref{fig:sampleEfficiency} shows evaluation score curves throughout the whole training process. This clearly proves the effectiveness of CCFDM compared to other methods.

\subsection{Discussion}

The SAC State is trained on hand-crafted features, considered the upper bound for performance and sample efficiency. However, according to Fig \ref{fig:sampleEfficiency}, CCFDM even outperforms SAC state in the Finger-Spin and Ball in Cup-Catch to some extent. It implies that hand-crafted (human-made) features are sometimes not the optimal features and thus learned features are potentially better features and produce better results without human effort.

The batch size has an important effect on our performance. During training and fine-tuning hyper-parameters, we realized that the larger batch size gives better performance. This is reasonable since CCFDM uses contrastive learning. However, a large batch size results in much lower training wall time. Thus, we should use it wisely. 

We tried to incorporate CURL \cite{srinivas2020curl} to CCFDM since doing so is quite straight forward. Interestingly, we found out that the performance remains the same while the training wall time increases due to additional computational cost. We argue that CCFDM contains all the CURL's effects. However, we need to work more to prove it with comprehensive experiments. The authors leave this task for future research.

\section{Conclusion}
\label{sec:majhead}
\footnotetext{Data and graph of baselines are referred from DrQ \cite{kostrikov2020image}}

This paper proposed a Curiosity Contrastive Forward Dynamics Model (CCFDM) framework, a contrastive reinforcement learning (RL) based framework that provides efficient image encoder (IE) learning for most RL algorithms. CCFDM showed that incorporating a forward dynamics model into contrastive learning framework helps IE capture both spatial and temporal information in order to improve IE learning. CCFDM also provides a smart exploration strategy based on FDM error which is simple but effective to encourage agents to better explore novel observations. CCFDM has proved effective in terms of improving sample efficiency and generalization. CCFDM is easy to incorporate with different modules to further improve RL performance. Future research should focus on developing more efficient and lightweight modules that can be integrated with CCFDM.


\bibliographystyle{IEEEtran}
\bibliography{main}

\end{document}